\documentclass[letterpaper, 10 pt, journal, twoside]{IEEEtran} 

\usepackage{graphics} 
\usepackage{epsfig}   
\usepackage{mathptmx} 
\usepackage{times}    
\usepackage{amsmath}  
\usepackage{amssymb}  
\usepackage{epstopdf}
\usepackage{hyperref}
\usepackage{pgfplots}
\usepackage{graphicx}
\usepackage{multirow}
\usepackage{xcolor}

\newcommand\ignore[1]{}

\usepackage{algorithm}
\usepackage{algpseudocode}
\algloopdefx{NoEndIf}[1]{\textbf{if} #1 \textbf{then}}

\usepackage{xspace}
\newcommand*{\eg}{\textit{e.g.}\@\xspace}
\newcommand*{\ie}{\textit{i.e.}\@\xspace}
\makeatletter
\newcommand*{\etc}{%
    \@ifnextchar{.}%
        {\textit{etc}}%
        {\textit{etc.}\@\xspace}%
}
\makeatother

\def\etal{\textit{et al.}~}

\def\cf{\emph{cf.},~}
\def\etal{\emph{et al.}~}

\def \path{\bp C}

\newlength\figureheight
\newlength\figurewidth

\graphicspath{ {./Figures/eps/} }

\title{Real-Time Dense Stereo Matching with ELAS on\\ FPGA Accelerated Embedded Devices}

\author{Oscar~Rahnama$^{1,2}$, Duncan~Frost$^{1}$, Ondrej~Miksik$^{1,3}$ and Philip H.S.~Torr$^{1}$%
\thanks{This work was supported by People Programme (Marie Curie Actions - ``Initial Training Networks'') of the EU FP7 under REA grant No. 317497 (EDISON), Technicolor, ERC grant ERC-2012-AdG 321162-HELIOS, EPSRC grant Seebibyte EP/M013774/1, EPSRC/MURI grant EP/N019474/1.}
\thanks{$^{1}$ Department of Engineering, University of Oxford, UK}
\thanks{$^{2}$ OxSight Ltd \quad $^{3}$ Emotech Labs}
}

\begin{document}
\maketitle

\begin{abstract}

For many applications in low-power real-time robotics, stereo cameras are the
sensors of choice for depth perception as they are typically cheaper and more
versatile than their active counterparts. Their biggest drawback, however, is
that they do not directly sense depth maps; instead, these must be
\emph{estimated} through data-intensive processes. Therefore, appropriate
algorithm selection plays an important role in achieving the desired performance
characteristics.

Motivated by applications in space and mobile robotics, we implement and
evaluate an FPGA-accelerated adaptation of the ELAS algorithm. Despite offering
one of the best trade-offs between efficiency and accuracy, ELAS has only been
shown to run at $1.5-3$ fps on a high-end CPU. Our system preserves all
intriguing properties of the original algorithm, such as the slanted plane
priors, but can achieve a frame rate of $47$fps whilst consuming under $4$W of
power. Unlike previous FPGA based designs, we take advantage of both components
on the CPU/FPGA System-on-Chip to showcase the strategy necessary to accelerate
more complex and computationally diverse algorithms for such low power,
real-time systems.
\end{abstract}


\section{Introduction}

\IEEEPARstart{I}{n} many areas of robotics, such as autonomous
navigation~\cite{oleynikova2015reactive, camellini20143dv,
Zainuddin2014AutonomousNO} and manipulation/grasping~\cite{lehnert2016sweet},
not only is the ability to perceive depth critical, but it needs to be obtained
very accurately and in real-time. On mobile or embedded platforms, power
consumption, cost, size and weight also become important factors to consider.
For instance, assistive augmented glasses should be mobile, lightweight and
ergonomic whilst retaining the ability to operate for long periods on limited
battery power \cite{hicks_depth-based_2013, miksik2015chi}.

Active methods of measuring depth, which are commonly used due to their high 
accuracy, carry certain disadvantages. LIDAR systems are often bulky, heavy and costly.
Infrared systems, on the other hand, are limited in their range, susceptible to
interference and, more importantly, constrained by ambient lighting. Passive
methods may not be limited by these factors, however, they are computationally
very expensive and their accuracy/latency depends heavily on the techniques
used.

Stereo matching algorithms can be broadly split into global energy minimization
methods and local matching techniques. The former are often more accurate, but
the generally large/irregular memory requirements and sequential/iterative
nature of their underlying algorithms make them dependent on powerful
processors for speed up, and even then, their frame rate is limited. Conversely,
local methods struggle with textureless and occluded regions, but the uniformity
of their computations and the absence of dependencies between pixels makes them
very suitable parallel acceleration.

Benefiting in part from the greater accessibility provided by CUDA/OpenCL, such
acceleration has been predominantly done with Graphics Processing Units (GPU).
However, Field Programmable Gate Arrays (FPGA) are becoming increasingly
competitive alternatives, especially in power limited systems, with their
capacity for in-stream processing, adherence to strict timings and
supremacy at sliding-window operations~\cite{fowers2012performance}.

Their effectiveness for stereo image processing has been previously
demonstrated~\cite{banz2010real, perez2016fpga, cocorullo2016efficient} with the
most accurate implementations usually relying on Semi-Global Matching
(SGM)~\cite{hirschmuller2005accurate}. However, as SGM is highly recursive,
memory intense and, in its entirety, ill suited to acceleration, those that do
either only partially implement it or sacrifice latency and throughput by
relying heavily on external memory.

In this paper, we investigate the adoption and acceleration of a competing
algorithm for low-power embedded systems. The algorithm, Efficient Large-Scale
Stereo (ELAS)~\cite{geiger2010efficient}, is the fastest CPU algorithm w.r.t.
resolution on the Middlebury dataset~\cite{ hirschmuller2007evaluation} and one
of the most accurate non-global methods.
ELAS is attractive since it very efficiently implements a slanted plane prior
while its dense depth estimation is fully decomposable over all pixels and,
hence, suitable for parallel processing.
Unfortunately, the intermediate step, \ie estimation of coarse scene geometry
through the triangulation of support points, is a very iterative, sequential and
conditional process with an unpredictable memory access pattern; making it
difficult to accelerate on an FPGA.

\begin{figure*}[!t]
  \centering
  \includegraphics[width=2\columnwidth]{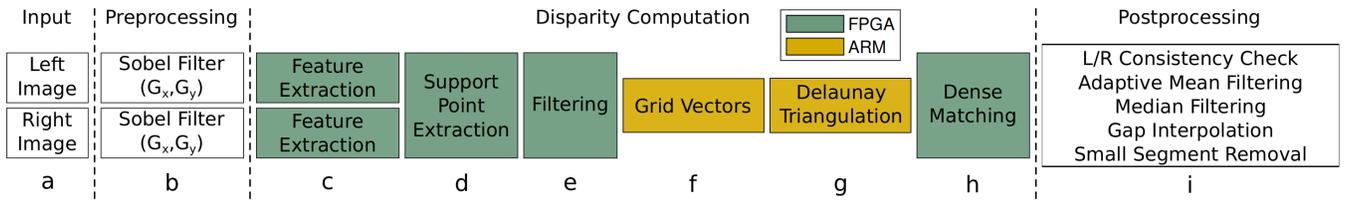}
  \vspace*{-2mm}
  \caption{ ELAS Overview: Extract a set of support points from gradient images
that are then used to establish priors for the dense matching stage. }
  \vspace*{-1.5mm}
  \label{fig:SystemHorizontal} 
\end{figure*}

To overcome this challenge, we propose the first stereo implementation which
collaboratively utilizes both components of an embedded CPU-FPGA System on Chip
(SoC) for the purpose of algorithm acceleration\footnote{Source code available
at \url{https://github.com/torrvision/ELAS_SoC}}. Other published
low-power systems achieve good frame rates by limiting the algorithms they
implement to those that can be fully processed by the FPGA, even when closely
coupled processors are available e.g.~\cite{camellini20143dv,
oleynikova2015reactive}. We, instead, seek to take advantage of both available
components to efficiently accelerate the more complicated/accurate ELAS
algorithm and demonstrate its feasibility for low-power systems. Accomplishing
this involves offloading the different stages of the processing pipeline onto
the component that best suits the computations involved. We discuss the rational
behind the chosen partitioning and explain why its the most suitable, describe
the key traits required in the design of efficient accelerators as well as the
changes made to best adapt the algorithm to the platform. Tested on both the
KITTI and New Tsukuba data sets, our system outperforms the frame rate of the
original by $\sim15-30\times$ with a rate of $47$fps ($1242 \times 375$ images)
and, in addition, improves upon its accuracy - all the while with under $4$W of
power consumption.

\section{Related work}

The pursuit of real-time stereo began in the
1980's~\cite{drumheller1986parallel}. Initial implementations of dense stereo
minimized relatively simple matching costs, \eg Sum of Absolute Differences
(SAD) or Sum of Squared Differences (SSD), between left and right image patches
evaluated along the epipolar lines. Kayaalp and Eckman~\cite{kayaalp1990near}
were one of the first to present such a system, capable of estimating disparity
over a $64$ disparity range in about one second for $256\times256$ images.

The first system capable of at least $30$ fps - on $200 \times 200$ images with
a 23 pixel disparity range - was on a custom platform built from off-the-shelf
components by Kanade~\etal\cite{kanade1995IROS, kanade1996CVPR}. Similarly to
\cite{kayaalp1990near}, they used a Sum of Sum of Absolute Difference (SSAD) but
rather than summing over the different color channels, they summed over the six
different cameras of their multi-camera system.

The first FPGA implementation of dense stereo matching used 16 Xilinx 4025
FPGAs~\cite{woodfill1997real}. It relied on the Census Transform (CT)~\cite{
zabih1994non} and computed 24 disparity levels over $320 \times 240$ images at
$42$ fps. Over about the next decade, FPGAs were repeatedly demonstrated as
suitable platforms for dense real-time stereo, however they mostly implemented
only variations of the SAD, SSAD, zero-mean SAD (ZSAD) and CT with additional
noise suppressing post-processing steps~\cite{hariyama2005fpga, perri2006sad,
cuadrado2006real, georgoulas2008real, ambrosch2010accurate}. Hence, the accuracy
of such approaches was not typically comparable with state-of-the-art models
which were formulated in global energy minimization
frameworks~\cite{hirschmuller2007evaluation, Scharstein2002,
scharstein2003high}.

Notable improvements in accuracy of FPGA implementations were made by
incorporating Semi-Global Matching (SGM)~\cite{hirschmuller2005accurate}. For
instance, Gehrig~\etal\cite{gehrig2009real} used a $3\times3$ window ZSAD along
a $64$ disparity range and minimized over $8$ separate directions to run at $27$
fps. Banz~\etal\cite{banz2010real} proposed a similar solution, but only
aggregated costs over $4$ directions with a rank transform~\cite{zabih1994non}.
SGM's larger scope managed to partly bridge the accuracy gap between strictly
local operators and global optimization methods whilst remaining suitable for
acceleration. However, SGM still has disadvantages such as large memory
requirements and a fronto-parallel bias. Alternative recent approaches have
shown some improvements in both frame rate and
accuracy~\cite{georgoulas2011real, werner2014hardware, wang2015real,
perez2015fpga, perez2016fpga, cocorullo2016efficient, li2016soc}, however, they
still lack the accuracy of SGM.

During the same period (2000-2010), graphics processing units (GPU) began to
appear as alternative platforms for algorithm acceleration~\cite{gong2005near,
lu2007fast,ernst2008mutual}. Although they offered speedup for sliding window
algorithms such as local stereo, GPUs typically under-performed and consumed more
power than their FPGA counterparts~\cite{fowers2012performance}. They were
therefore a less favorable option for truly embedded and real-time systems.

Currently, the fastest CPU stereo algorithm on the Middlebury
dataset~\cite{hirschmuller2007evaluation}, normalized w.r.t. resolution, is
Efficient Large-Area Stereo~\cite{geiger2010efficient}. It competes
with SGM in accuracy, but its diverse computational nature has it
overlooked in favor of other fully FPGA implementable algorithms. With new
closely coupled CPU/FPGA System-on-Chip devices, however, it stands to benefit a
lot in terms of acceleration.

\section{Preliminaries}
\label{sec:prelim}

\subsection{Original ELAS algorithm}

The Efficient Large-Scale Stereo Matching (ELAS)
algorithm~\cite{geiger2010efficient} relies on the assumption that not all
correspondences are \emph{equally difficult}. It first establishes a set of
sparse correspondences whose estimation is simpler and at the same time comes
with a higher degree of confidence. These correspondences provide a coarse
approximation of the scene geometry and are used to define a slanted plane prior
which guides the dense matching stage.

An overview of the ELAS is shown in Fig.~\ref{fig:SystemHorizontal}. To obtain the
set of sparse but confident correspondences (the ``support points''), the stereo
pair first passes through a SAD matching stage over the horizontal and vertical
gradients of the images, Fig.~\ref{fig:SystemHorizontal}(b-d). The resulting set
is sparse as only pixels with sufficiently unambiguous disparity values are
kept. This criterion is measured by comparing the distance between the first and
second minima of the SAD evaluations across the disparity range. The results
from this stage then undergo a further filtering procedure,
Fig.~\ref{fig:SystemHorizontal}(e), to remove implausible and redundant values
which would respectively corrupt or unnecessarily complicate the coarse 3D
representation. This filtering process compares the sparse values to neighbors
within a window region to ensure that they are consistent and removes identical
values along the same row or column.

\begin{figure}[!t] \centering
  \includegraphics[scale=0.40, height=3.4cm]{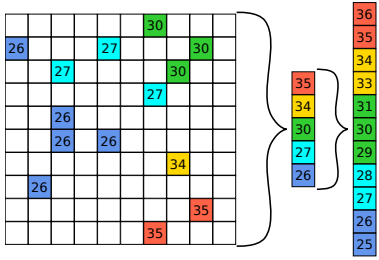} 
  \caption{Pooling support points within a sub-region to create a grid vector.
    The sparse set of correspondences in every given grid-region of an
    image are pooled together to create a characteristic search vector. This
    search vector is expanded to include immediate neighbors of included
    support points ($\pm 1$)}
  \label{fig:SubRegions}
\end{figure}

The set of support points is then used to guide the dense stereo matching stage
(Fig.~\ref{fig:SystemHorizontal}(h)) in two separate ways. First, the set of
support points is used to define a slanted plane prior which guides the dense
matching stage. To this end, ELAS uses Delaunay triangulation to construct a
mesh which approximates coarse scene geometry
(Fig.~\ref{fig:SystemHorizontal}(g)). Second, this slanted plane prior is used
to limit the disparity values evaluated during the dense matching stage. This
range is expanded to include immediately neighboring values ($\pm 1$),
which gives an algorithm a chance to recover in case the initial mesh is
incorrect (Fig.~\ref{fig:SubRegions} and Fig.~\ref{fig:SystemHorizontal}(f)).

Following the dense matching stage, ELAS uses post-processing
(Fig.~\ref{fig:SystemHorizontal}(i)) to invalidate occluded pixels and
further improves smoothness across the image. Although post-processing plays
an important role in obtaining an accurate final result, it is not a core part
of the algorithm or unique to it, and therefore ignored. 

\subsection{Platform}
\label{sec:platform}

We use Xilinx's ZC706 development board with the XC7Z045 SoC; this is a
heterogeneous chip that incorporates an ARM Cortex A9 processor operating
at 800MHz and a 28nm Kintex series FPGA on the same die. The collocation of the
two components ultimately serves to increase the overall throughput of the
system as it allows for rapid and efficient exchange of data. The resources
available on the FPGA include: 218600 Look Up Tables (LUT), 437200 Flip Flops
(FF), 900 DSP48 Blocks, 1090 18K Block RAMs (BRAM).

\subsection{High level synthesis} 

Vivado High Level Synthesis (VHLS) provides a higher abstraction approach to
FPGA block implementation by synthesizing designs described in a high level
language such a C/C++ into equivalent hardware descriptions. A deep
understanding of the underlying hardware architecture is still required, but it
alleviates the burden of adopting a low level hardware description languages
such as VHLD/Verilog. By abstracting fine grained, less critical details of the
design, VHLS accelerates and facilitates development with FPGAs. Although
accelerators designed with this higher abstraction approach may not be as
optimized or as resource efficient as those designed with low level hardware
description languages, the ability to deploy, modify and test them much more
rapidly is a reasonable compromise. Hence, we implemented all accelerators using
VHLS.

\section{System Overview}

\begin{figure}[!t]
  \centering
  \includegraphics[width=0.839\columnwidth]{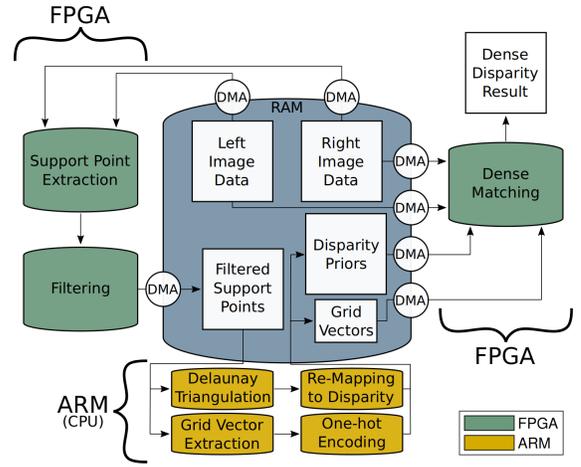} 
  \caption{Overview of System-on-Chip (SoC) implementation. Compute-intensive
    tasks are offloaded to FPGA accelerators whereas conditional/sequential 
    tasks are handled by ARM CPU. Communication between CPU and FPGA
    is handled by Direct Memory Access blocks in the FPGA.}
  \label{fig:System}
\end{figure}

Fig.~\ref{fig:System} shows the overall system implemented on Zynq SoC platform.
Determining which parts of the algorithm are offloaded onto dedicated
accelerators and which are proceed on the ARM CPU is a twofold process. First,
the entire algorithm's CPU runtime is profiled to get an estimate of the time
spent in each function. The main bottlenecks are then identified and the
algorithm is mapped and broken down into its main components. In the subsequent
step, the computational nature of the different components are evaluated and
matched to the most appropriate component for processing.

\paragraph{FPGA}
FPGA accelerators are severely hampered if they require communicating with
external memory or if they contain many divergent datapaths through them.
However, they excel at performing a variety of operations,
simultaneously, on a large range of data. As such, functions that process blocks
of data with well defined, relatively local, memory access patterns and limited
amounts of conditional branching can benefit tremendously from such
acceleration.

In ELAS, the functions responsible for support point extraction, filtering and
block matching fit such criteria. Therefore, as denoted by the green blocks in
Fig.~\ref{fig:System}, these are offloaded onto dedicated FPGA accelerators.
These accelerators can either have data
transferred in-between them directly (Sparse $\rightarrow$ Filtering) or to and
from the the RAM through Direct Memory Access (DMA) blocks in the FPGA
(programmed by the CPU). The accelerators process the data \emph{in-stream} -
only storing small portions of the overall stream (\cf
Sec.~\ref{sec:ImplDetails}) and outputting data at the same rate at which it is
received. This processing style best compliments the raster pixel readout of
modern image sensors and allows for top level pipelining in between successive
FPGA accelerators.

\paragraph{ARM CPU}
Functions with very unpredictable memory access patterns as
well as those with a high amount of conditional branching are very ill suited
for FPGA acceleration. These, instead, benefit more from the ARM CPU's faster
processing speeds, its sequential processing style (invariant to branching) and
its equal, but longer, access to memory (disregarding cache hit/misses).

As denoted by the yellow blocks in Fig.~\ref{fig:System}, the ELAS processes
that are handled in this manner are the Delaunay triangulation as well as the
remapping of the slanted plane priors into disparity priors. Grid vector
extraction and one-hot encoding (\cf Sec.~\ref{sec:sec6}) are also done on the
ARM. Although grid vector extraction would appear to be a candidate for FPGA
acceleration as it ``pools'' values within a local memory region with a window
like operation, as shown in Fig.~\ref{fig:SubRegions}, in reality, as it only
operates on a single value at a time, it benefits more from the faster clock of
ARM.


\section{Key Accelerator Design Traits}
\label{sec:ImplDetails}

\subsection{FPGA memory management}

\begin{figure}[!t]
  \centering
  \includegraphics[height=4.5cm,scale=0.5]{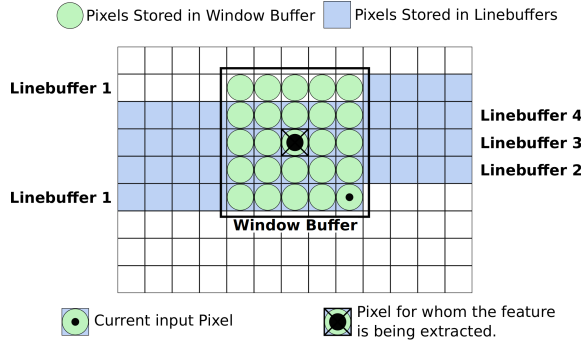}
  \caption{Memory requirements for sliding window operations in FPGA
    accelerators. Line buffers (blue) are used to store large amounts of data
    but can used to provide a single value per clock cycle. Window buffers
    (green) are local registers used to store immediately required data.}
  \label{fig:Extractbuffers}
\end{figure}

To achieve the required parallelism within the FPGA blocks, all the necessary
data for a set of computations must be available on the same clock cycle as
those computations are to occur. Memory management, therefore, has the largest
impact on accelerator throughput and latency. Fig.~\ref{fig:Extractbuffers}
shows the combination of block RAMs and local memory that are used as the core
components for this purpose.
Block RAMs (BRAMs) are the most resource efficient stores of large quantities of
data. In the accelerators, they behave as line buffers, storing previously
received pixel information into the FPGA fabric. For a given $W \times W$ matching
window, each image's pixel data is collected by a set of \(2\times W\) line
buffers. 
On every clock cycle, each line buffer shifts its contents at a given index to
the line buffer above it and a new pixel is read into the free space created at
that index of the bottom-most buffer. Similarly, the data that was in the
topmost buffer is shifted out as it is no longer required - only a
fraction of the overall image data is ever held in the FPGA.

Each BRAM, however, is only able to read and write one value per clock cycle.
Therefore, a set of line buffers only supplies one column of pixel data per
clock cycle. Most computations in the accelerators, however, operate over
\emph{windows} of pixels and therefore this alone is insufficient in meeting the
memory requirements for a high throughput/low latency design. Instead, 
an additional \(W \times W\) size ``window'' buffer is necessary (Fig.~\ref{fig:Extractbuffers}).
As it consists
of local registers within the FPGA block which are all instantaneously
accessible, this is resource expensive. On every clock cycle, the contents of the window are updated by
shifting all columns once to the left. The rightmost column is read in from the
values stored in the line buffers (including the latest pixel value).

By combining the use of storage elements in this manner, we efficiently achieve
access to all the necessary data on the same clock cycle on which it is used. No
additional clock cycles need to be spent on memory access.

\subsection{Pipelining}

Although VHLS handles timing considerations and data flow control of FPGA
accelerators, the throughput and latency it achieves depends on the propagation
delays within the accelerators as well as the desired amount of pipelining and
overall clock frequency.

When maximizing the throughput of an accelerator, pipelining is necessary when
its total internal propagation delay exceeds the clock period to which it is
constrained (pixel read in/out rate). By introducing pipelining, the accelerator
is able to meet the clock frequency constraint by dividing and spreading its
operation over multiple clock cycles. Each sub-stage is separated from the next
with flip flops that store intermediate values and therefore pipelining improves
utilization. Ultimately, however, it results in the accelerator being shared
across a set of inputs as each sub-stage processes a new input on every clock
cycle - a larger amount of data is being simultaneously acted upon.

Other than the flip flop requirement, the trade-off with pipelining is that the
number of clock cycles between first input and output increases by the amount of
pipeline stages and the propagation delay experienced by a single pixel is
longer as each sub-stage's delay is extended to that of the longest sub-stage.
In image processing, however, as large quantities of pixel data pass through the
accelerators, the latency introduced by pipelining is not only negligible, but
significantly outweighed by the ability to output data at a much faster rate. It
plays a significant role in achieving the desired frame rate in our design.

\section{Platform Conscious Algorithm Changes}
\label{sec:sec6}

\subsection{Feature selection}

Efficient implementation of original ELAS uses SIMD accelerators with fixed
widths of $16$ bytes for feature extraction and matching. Such an
implementation, however, lacks flexibility since the number of pixels it can
process is limited and must be in multiples of 16 (for 8 bits). The result is
that for a given $W \times W$ window, only a subset of pixels contained within
it are used for matching purposes.

Due to our memory management, all pixel data within a window is available and
therefore no speed penalty is incurred by using it
(Fig~\ref{fig:Extractbuffers}). It also improves the accuracy since we use a
larger number of pixels in the matching process. We use the Census Transform
descriptors with Hamming Distance instead of the SAD as it achieves illumination
invariant matching without the need for an additional pre-processing step (Sobel
filter Fig.~\ref{fig:SystemHorizontal}(b)).

\begin{figure}[!t]
  \centering
  \includegraphics[width=0.8\columnwidth]{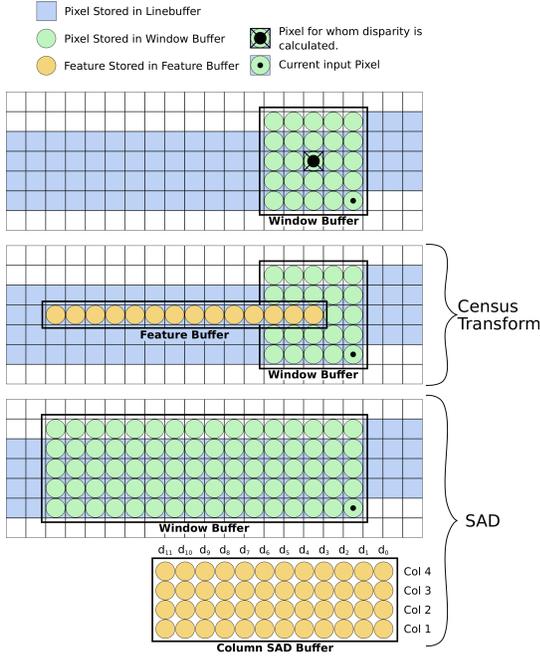}
  \caption{Comparison of Extraction and Matching Implementation with Census
    Transform and Sum of Absolute Differences}
  \label{fig:Matchbuffers}
\end{figure}

Even if the pre-processing is discounted, the SAD generally requires more
resources than the CT. As shown in Fig.~\ref{fig:Matchbuffers}, unlike the CT
that reuses previously extracted features, the SADs must be recomputed every
time. Also, SAD implementations that achieve similar throughput, such as the one
in~\cite{rahnama2017embedded}, require an additional window buffer to store
previous column SAD computations (bottom of Fig.~\ref{fig:Matchbuffers}). Thus
the resource requirement is greater due to both the greater number of
computations as well as the greater need for memory.

These changes result in a feature descriptor that is shorter in bit length
($81/25$ compared to the original $512/128$), while containing information about
larger pixel neighborhood than ELAS's CPU implementation.

\subsection{Measuring ambiguity}

The support point extraction is done slightly differently to the original
algorithm. We replace the original criteria which assumes a match is unambiguous
if
\begin{equation}
  \frac{m_1}{m_2} \leq 0.9,
\end{equation}
where $m_1$ and $m_2$ are the first and second mimima respectively. This is
equivalent to thresholding $m_1$
\begin{equation}
  m_1 \leq 0.9 m_2 = T_{err}(m_2),
\end{equation}
However, implementing such comparison on an FPGA requires a number of DSP
blocks. Hence we approximate the threshold $T_{err}(m_2)$ with a shift-sum
\begin{equation}
    T_{err}(m_2) = 0.9m_2 \approx \frac{m_2}{2} + \frac{m_2}{4} + \frac{m_2}{8} + \frac{m_2}{32} = 0.90625m_2.
\end{equation}

This eliminates the need for DSP blocks as shift-summing is fully accomplished
within the LUT fabric of the FPGA.

\subsection{Filtering support points}

\begin{figure}[!t]
  \centering
  \includegraphics[width=0.8\columnwidth]{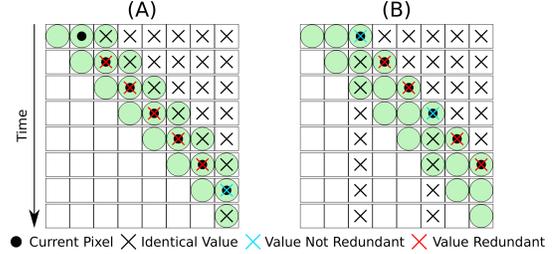} 
  \caption{One dimensional simplification of redundancy verification (A)
   Searching both forwards and backwards propagates redundancy into the value's
   non-existence (B) Searching strictly backwards retains the value at the desired
   frequency}
  \label{fig:redun}
\end{figure}

The original algorithm uses both past and future values in the data stream for
redundancy check. As shown in the simplified 1-D example of
Fig.~\ref{fig:redun}(A), the shared values used to flag redundancy are often
made redundant by instances further ahead. Instead, the ``filter'' FPGA
accelerator only relies on past values when determining redundancy. This ensures
the shared values are less frequent rather than non-existent.

\subsection{Data Reduction}

To accommodate for the static bit-width of accelerator ports and to minimize the
data transferred to the FPGA, we introduce new data reduction steps to ELAS (on
the ARM). Referring to Fig.~\ref{fig:System}, the first one-hot encodes the grid vectors
from a variable length byte array into a statically sized bitwise
representation. The second converts the result of the Delaunay Triangulation
from a variable mesh of triangles into a static, input image sized, matrix of
disparity priors. 

\section{Experimental Results}

To demonstrate the effectiveness of our approach, we provide a detailed
evaluation across differently parameterized sets of implementations and evaluate
them on both the KITTI and New Tsukuba
data sets.

As FPGA accelerators can not be easily reconfigured for different
image resolutions during testing, we only use $310$ of the $400$ KITTI
image pairs that have the same $1242 \times 375$ resolution. The New
Tsukuba consists of 1800 images with resolution of $640 \times 480$,
of which we use the provided subset of 200 image-pairs.

\subsection{Accuracy}

We begin by verifying how the accuracy of the FPGA accelerated version of ELAS,
following the modifications made to adapt it to the SoC platform,
compares to that of the original algorithm. To this end, we use the standard
accuracy metric from the KITTI benchmark which measures the relative number of
estimated disparities which differ from ground truth by both an absolute amount
of at least 3 and a relative one of at least 5\%.

To ensure a fair comparison, the number of support points used to establish the
prior should be approximately the same. As not all pixels are considered for
support point extraction in the original algorithm, we find that this occurs
when we use \(\frac{1}{14}\)-th of the number of total extracted support points
in our method. We also use the same window sizes for both matching stages, \ie
$9 \times 9$ and $5 \times 5$, respectively. As previously explained, accuracy
is measured without post-processing/refinement as these processes aren't unique
to ELAS and are more susceptible to dataset ``fine tuning''. With these
parameters, the original implementation tested without post-processing over the
same image set, obtains an average error of $17.9\%$ while our implementation
achieves an improved $16.5\%$. With many other configurations tested,
Table~\ref{wintable}, we find that the embedded version is generally more
accurate, except when window sizes are made too small or when support points are
overly sub-sampled. Unlike the hard-coded $9 \times 9$ and $5
\times 5$ windows of the original, we are able to quickly configure different
window sizes to vary accuracy without any impact on frame-rate. As larger
windows provide more information in matching, they also result in more accurate
depth maps. However this only holds up to a certain size; eventually the
inherent fronto-parallel bias of square matching windows begins negatively
impact results.

\begin{table*}[]
\centering
\caption{Impact of Window Sizes used during matching on frame rate, accuracy and
resource utilization}
\label{wintable}
\begin{tabular}{|c|c|c|c|c||c|c||c|c||c|c|c|c|c|}
\hline
\multirow{2}{*}{Data Set} & \multirow{2}{*}{\begin{tabular}[c]{@{}c@{}}Window\\ Size\end{tabular}} & \multirow{2}{*}{\begin{tabular}[c]{@{}c@{}}Window\\ Size\end{tabular}} & \multicolumn{2}{c||}{CPU ELAS} & \multicolumn{2}{c||}{1/8} & \multicolumn{2}{c||}{1/32}  & \multicolumn{3}{c|}{Resource Utilization [\%]} & \multicolumn{2}{c|}{Power (Watts)} \\ \cline{4-14} 
 &  &  & FPS & Error \% & FPS & Error \% & FPS & Error \% & LUT & FF & BRAM & ARM & FPGA \\ \hline
\multirow{12}{*}{\begin{tabular}[c]{@{}c@{}}KITTI \\ \(1242 \times 375\)\end{tabular}} 
& \multirow{3}{*}{\(7 \times 7\)} & \(3 \times 3\)& \multirow{12}{*}{1.5 - 3} & \multirow{12}{*}{17.9} 
    & 17.3 & 18.4 & 29.2 & 23.2 & 22.0 & 14.5 & 10.5 & 1.70 & 0.91\\  \cline{3-3} \cline{6-14} 
 &  &\(5 \times 5\)& & & 17.4 & 16.5 & 29.2 & 21.7 & 24.2 & 15.3 & 10.8 & 1.70 & 1.06\\  \cline{3-3} \cline{6-14} 
 &  &\(7 \times 7\)& & & 17.4 & 15.8 & 29.2 & 21.4 & 27.2 & 16.7 & 11.2 & 1.70 & 1.08\\  \cline{2-3} \cline{6-14} 
 & \multirow{3}{*}{\(9 \times 9\)}                                                  
    &\(3 \times 3\)& & & 12.5 & 17.2 & 24.5 & 20.0 & 26.2 & 17.1 & 10.8 & 1.70 & 1.00\\  \cline{3-3} \cline{6-14} 
 &  &\(5 \times 5\)& & & 12.4 & 14.5 & 24.5 & 17.7 & 28.5 & 17.9 & 11.2 & 1.70 & 1.17\\  \cline{3-3} \cline{6-14} 
 &  &\(7 \times 7\)& & & 12.3 & 13.7 & 24.3 & 17.0 & 31.7 & 19.3 & 11.6 & 1.70 & 1.16\\  \cline{2-3} \cline{6-14} 
 & \multirow{3}{*}{\(11 \times 11\)}                                                
    &\(3 \times 3\)& & & 10.5 & 17.6 & 22.3 & 19.5 & 32.0 & 20.4 & 11.2 & 1.70 & 1.08\\  \cline{3-3} \cline{6-14} 
 &  &\(5 \times 5\)& & & 10.5 & 14.5 & 22.3 & 16.6 & 34.0 & 21.2 & 11.6 & 1.70 & 1.20\\  \cline{3-3} \cline{6-14} 
 &  &\(7 \times 7\)& & & 10.5 & 13.6 & 22.3 & 15.7 & 37.3 & 22.6 & 11.9 & 1.70 & 1.21\\  \cline{2-3} \cline{6-14} 
 & \multirow{3}{*}{\(13 \times 13\)}                                                
    &\(3 \times 3\)& & & 9.4  & 18.2 & 21.0 & 19.4 & 38.5 & 24.2 & 11.6 & 1.70 & 1.12\\  \cline{3-3} \cline{6-14}
 &  &\(5 \times 5\)& & & 9.5  & 14.9 & 21.0 & 16.3 & 40.4 & 25.1 & 11.9 & 1.70 & 1.27\\  \cline{3-3} \cline{6-14}
 &  &\(7 \times 7\)& & & 9.5  & 13.9 & 21.0 & 15.4 & 43.5 & 26.5 & 12.3 & 1.70 & 1.28\\  \hline
 \multirow{2}{*}{\begin{tabular}[c]{@{}c@{}}Tsukuba \\ \(640 \times 480\)\end{tabular}}
 & \(9 \times 9\)  & \(5 \times 5\) & \multirow{2}{*}{N/A} & \multirow{2}{*}{6.4} & 17.6 & 6.8 & 36.2 & 6.4 & 28.1 & 17.0 & 9.7  & 1.70 & 1.00 \\ \cline{2-3} \cline{6-14} 
 & \(11 \times 11\)& \(7 \times 7\) & & & 14.8 & 6.4 & 32.9 & 5.9 & 37.4 & 21.7 & 10.5 & 1.70 & 1.23 \\ \hline
\end{tabular}
\end{table*}

\subsection{Per-frame processing time}

\begin{figure}[!t]
  \centering
  \includegraphics[width=0.95\columnwidth]{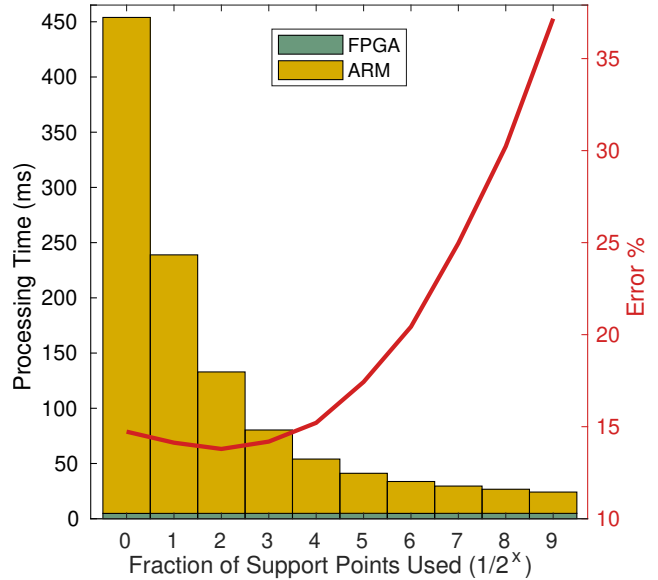} 
  \caption{Processing time and related accuracy w.r.t down sampling}
  \label{fig:Graph1}
\end{figure}

Fig.~\ref{fig:Graph1} (left axes) illustrates the proportion of the overall
processing time of the ARM against that of the FPGA. The FPGA portion is
inclusive of the time spent transferring data by the DMAs. Additionally, the
results are reported across different support point densities which we regulate
through down sampling.
As shown, although the time spent on the ARM is proportional to the number of
support points used and shortens significantly with down sampling, it
nonetheless dominates the overall processing time.

In contrast, the combined processing time of the FPGA accelerators is mostly
unaffected by changes in parameters such as matching window size, disparity
range or the number of support points. As they have a constant throughput of $1$
pixel/clock cycle, their processing time is, instead, predominantly a function
image resolution. On average, it takes only \(4.84\pm.02\)ms for the KITTI
images ($1242 \times 375$px) and \(3.19\pm.02\)ms on the New Tsukuba
($640\times480$px). The $1.51\times$ difference corresponds exactly to the pixel
ratio difference.

The line plot in Fig.~\ref{fig:Graph1} (right axes) shows the error percentage vs.
the number of support points controlled by down-sampling. Whilst slightly
unintuitive, the best accuracy is not achieved with the largest number of
support points (peak at $\frac{1}{8}$ of the support points). This is likely due
to the noisy nature of sparse correspondence matching. Following this peak,
accuracy gradually decreases with reduced number of support points as the
resulting planar surfaces become less accurate coarse approximations of scene
geometry.

Interestingly, although the FPGA accelerators run at constant time, the matching
window size of the support point extraction stage is negatively correlated with
overall frame rate. Larger windows do incur a greater initial latency to account
for the additionally used line buffers, but this difference is negligible
(evidenced by invariance of the frame rate to the window size of dense matching)
and can not account for this difference. Instead, the frame rate reduction is
actually due to the increased number of support points resulting from the
extraction using larger matching windows - this impacts the ARM's workload.
Therefore, the main bottleneck is the Delaunay Triangulation which is in stark
contrast to the one reported in the original CPU implementation.

\subsection{Power and resource consumption}

The high throughput capability of accelerators is limited by the number of
circuit elements available within the FPGA fabric. In order to report this
``resource utilization'' (Table~\ref{wintable}), we split it into the three main
types of blocks (we exclude DSP blocks as they are negligibly used). As
expected, the LUTs, used for the combinational logic and instantaneous memory,
are the most predominantly utilized and this amount depends strongly on the
window size of dense matching. The flip flops, utilized primarily for
pipelining, share a similar dependency, but to a lesser extent. The BRAMs, used
as line buffers are purely a function of the cumulative image rows required for
a given set of windows.

One of the most important advantages of the proposed implementation is the power
efficient computing that it enables (\cf last two columns of
Table~\ref{wintable}). The ARM processor, running at a steady 800 Mhz, accounts
for a constant but majority share of the power. In contrast, the power consumed
by the FPGA is much more controlled and directly proportional to the portion of
FPGA logic that is being utilized. Altogether, however, the results highlight
that the implementation is not only capable of running the algorithm in
real-time, it succeeds in doing this with under $3$W of power (in contrast to
powerful desktop CPUs which typically require $>100$W).

\subsection{Throughput}

To further increase frame rate, we explore operating over multiple images simultaneously by taking
advantage of the additional core of the ARM and separate but identical accelerators in the FPGA.
This effectively doubles the system and therefore the resources and power used by the FPGA double
(minus some shared overhead). Conversely, the ARM's power consumption remains the same, at $1.7$W.
For example, a dual system with $(9 \times 9)$ and $(5 \times 5)$ matching windows utilizes $56.8\%$
of LUTs, $35\%$ of FFs, $22.8\%$ of BRAMs and consumes $3.67$W. The most accurate configuration
using this multi-threaded/multi-accelerator approach runs at $23.7$fps ($3.74$W). A faster version, whose
accuracy is poorer but still better than the original CPU versions runs at
$47.0$fps ($3.67$W).

\section{Comparison \& Discussion}

\begin{table}[]
\centering
\caption{Time Breakdowns}
\label{timetable}
\begin{tabular}{|c|c|c|c|}
\hline
Cones (900 x 750) & \begin{tabular}[c]{@{}c@{}}Time (ms)\\ i7-only (orig.)\end{tabular} & \begin{tabular}[c]{@{}c@{}}Time (ms)\\ ARM+FPGA\end{tabular} & \begin{tabular}[c]{@{}c@{}}Time (ms)\\ i7+FPGA\end{tabular} \\ \hline
Support Points & 118 & 3.5 & 3.5 \\ \hline
Triangulation  & 7   & 84.42 & 7 \\ \hline
Matching       & 359 & 3.5 & 3.5 \\ \hline
\end{tabular}
\end{table}

Comparing these results to what was achieved in the original paper, it is clear
that parallelizing key parts of the algorithm has successfully led to
significantly faster - up to $30\times$ - real-time frame-rates. Despite the
achievement, the results also reveal some of ELAS's weaknesses for power-limited
platforms. From Table~\ref{timetable}, where the time breakdown for each stage
is compared across systems, we see how the SoC manages $~100\times$ throughput
increase for both matching stages even though it processes more data, i.e. all
pixels considered for support point and full matching windows. However, with the
low-power CPU paling in performance compared to its desktop counterpart, the
triangulation procedure - seemingly insignificant in the original paper - is
$>12$ times longer and dominates the processing time on the SoC. Thus, although
ELAS is exemplified as one of the fastest stereo algorithm, its dependence on a
very sequential procedure makes it also dependent on a powerful processor to
achieve maximal speed up. In the last column of Table~\ref{timetable}, we show
the processing times which one could obtain if an SoC, combining the same FPGA
with an Intel Core i7 CPU instead of the ARM, were used to accelerate the
algorithm with the same proposed approach. On such a platform, the frame rate of
ELAS exceeds $70$fps, but power consumption would also exceed $100$W.

In terms of accuracy, the improvements over the original can be attributed to
the full matching windows/CT as opposed to the randomly sub-sampled SAD of the
original. These sub-sampled windows were needed in the original to
speed up the CPU processing time. In contrast, with the FPGA accelerators, not
only is window matching speed independent of the number of pixels considered,
but full windows result in a more efficient use of resources. As a corollary,
unlike CPU ELAS whose runtime is coupled to its tailored windows, our embedded
version can accommodate various window sizes and disparity ranges without being
concerned about the impact on latency.

We compare the performance of our system to the fastest implementation currently
reported on the KITTI benchmark, referred to as ``CSCT+SGM+MF''
(CSM)~\cite{hernandez2016embedded} and which, at its core, implements SGM - the
competing algorithm in embedded, real-time systems. It reports an 8.24\% error
rate at 156 fps on a 250W NVIDIA Titan X. As CSM's result incorporates
smoothing/refinement both inherently through SGM and through an additional
median filter, we pass the results of our most accurate configuration through a
median filter for the sake of comparison. With only this one additional
post-processing step, we obtain a new error rate of 9.52\% - already slightly
better than the accuracy ELAS reports on the KITTI benchmark following all
post-processing/refinement. Although CSM may still be marginally more accurate
with a faster frame rate, it requires substantially more power at 250W. This is
equivalent to a per Watt frame rate of $0.62$fps/W. In their original paper, the
authors also attempt a more power efficient implementation (same accuracy) on a
mobile NVIDIA Tegra X1 GPU (10W). After scaling their results to the KITTI
resolution, this more power efficient system manages only 13.8fps with a
resulting efficiency of $1.38$fps/W. Our system, in contrast, offers a $23.7$fps
frame rate with a corresponding $6.34$fps/W efficiency. This is a $~10\times$
improvement over the faster Titan X system and $~5\times$ over the slower Tegra
X1 one.

The FPGA accelerators in this system were described in C++ and then converted
into logic with VHLS. In our experience, although the tool did eliminate the
need for writing low-level VHDL/Verilog, it still relied heavily on the user's
deep knowledge of the target circuit. The original algorithm had to be
completely re-engineered to comply with the hardware framework. As well as
motivating the previously described modifications, this included adhering to a
regular, timing-strict processing chain, minimizing any inter-process
dependencies and eliminating conditional operations/branches/exceptions.

\section{Conclusion}

In this work, we have disassembled and reconstructed the ELAS algorithm onto an
ARM + FPGA SoC with the purpose of evaluating its suitability for low-power,
real-time embedded systems. By taking advantage of the immense parallelism
available with FPGAs and by better adapting the algorithm for it, we not only
successfully accelerate the frame rate by up to $30\times$, but we also
demonstrate an improvement in accuracy. All this is achieved with under 4W of
power which makes it $5-10$ more efficient, on a frame rate per Watt basis, than
competing algorithms on KITTI.

Through the iterative process of adapting the algorithm to the platform as well
the starkly different resulting processing time breakdown we obtained,
fundamental principles were gleaned for the future design of accurate, but
ultimately real-world applicable, algorithms. Specifically, with parallelism
being of paramount importance, any strictly sequential or iterative processes
must be kept to a minimum as these will cause severe bottlenecks. Their
acceleration depends on faster processors, and as CPU frequency is directly
proportional to power consumption, this quickly incurs greater power
requirements that are unrealistic in space, aerial or mobile robotics.
Conversely, accelerators excel at simultaneously processing vast amounts of data
as long as it is available and effectively managed in the fabric of the FPGA.
Therefore, compromises that may have made sense for a strictly CPU system, such
as sacrificing accuracy for speed by computing with fewer pixels, are no longer
necessary and should be entirely avoided.

Finally, although the newly developed high-level design tools by Xilinx do
indeed facilitate the access, speed and and transportability of designing on
FPGAs, one must be still possess a strong understanding of hardware design in
order to efficiently implement accelerators with them.

\bibliography{new_references}
\bibliographystyle{IEEEtran}
\end{document}